\pdfoutput=1

\documentclass[11pt]{article}

\usepackage[]{EMNLP2022}

\usepackage{times}
\usepackage{latexsym}
\usepackage{multirow}

\usepackage[T1]{fontenc}

\usepackage[utf8]{inputenc}

\usepackage{microtype}
\usepackage{amsfonts}

\usepackage{inconsolata}
\usepackage{microtype}
\usepackage{amsmath}
\usepackage{algorithm}
\usepackage{algorithmicx}
\usepackage{algpseudocode}
\usepackage{booktabs}
\usepackage{pgfplots} 
\usepackage{subcaption}
\usepackage{tikz}

\usepackage{graphicx}

\makeatletter
\newcommand*\bigcdot{\mathpalette\bigcdot@{.5}}
\newcommand*\bigcdot@[2]{\mathbin{\vcenter{\hbox{\scalebox{#2}{$\m@th#1\bullet$}}}}}
\makeatother
\title{NoPPA: Non-Parametric Pairwise Attention Random Walk Model\\ for Sentence Representation}

\author{Xuansheng Wu \\
  University of Georgia \\
  210 S Jackson Street, Athens \\
  Georgia, United States \\
  \texttt{Xuansheng.Wu@uga.edu} \\\And
  Zhiyi Zhao \\
  Tufts University \\
  161 College Avenue, Medford \\
  Oregon, United States \\ 
  \texttt{zhiyi.zhao@tufts.edu} \\ \And
  Ninghao Liu \\
  University of Georgia \\
  210 S Jackson Street, Athens \\
  Georgia, United States \\
  \texttt{Ninghao.Liu@uga.edu}
  }

\begin{document}
\maketitle
\begin{abstract}
We propose a novel non-parametric/un-trainable language model, named Non-Parametric Pairwise Attention Random Walk Model (NoPPA), to generate sentence embedding only with pre-trained word embedding and pre-counted word frequency. To the best we know, this study is the first successful attempt to break the constraint on bag-of-words assumption with a non-parametric attention mechanism. We evaluate our method on eight different downstream classification tasks. The experiment results show that NoPPA outperforms all kinds of bag-of-words-based methods in each dataset and provides a comparable or better performance than the state-of-the-art non-parametric methods on average. Furthermore, visualization supports that NoPPA can understand contextual topics, common phrases, and word causalities. 
Our model is available at https://github.com/JacksonWuxs/NoPPA.
\end{abstract}

\section{Introduction}
Precisely representing sentence-level semantic information is a cornerstone in widely natural language understanding tasks. In the era of pre-trained language models \citep{bert, floridi2020gpt}, researchers proposed various training strategies \citep{reimers2019sentence, barkan2020scalable, wu2021smoothed, cheng2021dual, jiang2022promptbert}, post-processing procedures \citep{li2020sentence, huang2021whiteningbert}, and pooling methods \citep{reimers2019sentence, wang2020sbert} to generate high-quality sentence embedding from such large-scale models. However, the large numbers of parameters inside pre-trained models require huge computation recourse, which might not always be accessible in all scenarios. 

To develop low-resource models, some researchers choose another direction to encode sentence embedding by using simpler language models and pre-trained word embedding \citep{mikolov2013efficient, pennington2014glove, wieting2015paraphrase}. The most straight forward way in this direction is weighted averaging word embedding \citep{wieting2015towards, arora2017simple, ethayarajh2018unsupervised}, which essentially is a bag-of-words model under an independent assumption of words. These methods are simple, while their performance suffers from bag-of-words assumption seriously. Enhancing word embedding by combining multi-resources word embedding \citep{mekala2016scdv, ruckle2018concatenated} was a popular strategy to improve bag-of-words models in the early days. Recent studies usually remove the bag-of-words assumption to improve such simple language model by capturing time information \citep{kayal2019eigensent, almarwani2019efficient} and using semantic subspace analysis \citep{ionescu2019vector, wang2021efficient}.

In this paper, we propose the Non-Parametric Pairwise Attentive Random Walk Model (NoPPA), which boosts the modeling level of bag-of-words models from span N-grams to pairwise Bi-grams with non-parametric attention. Generally, NoPPA estimates the probability of a word appearing in a given sentence based on the non-contextual probability of words, the fluency of the word in the current sentence, and the likelihood that the word expresses the intent of the sentence. We further prove that the estimation of sentence embedding coming from NoPPA is weighted averaged pairwise word embedding. Specifically, we calculate the pairwise word embedding by a non-linear activation function over the difference of the word pairs. We evaluate our method on eight different text classification tasks. Results show that NoPPA outperforms all weighted-averaging-based methods and most of state-of-the-arts non-parametric methods, including time-information-infused and latent-space-analysis-based methods. Visual analysis shows that NoPPA dynamically adjusts the contributions of words to sentence embedding respecting different contexts, while the proposed non-parametric pairwise attention captures common multi-word phrases and causation between words.

We organize this paper as follows. In Section~\ref{section2}, we first reviews some recent studies in sentence embedding, including parametric and non-parametric methods. Then, we introduce our method in Section~\ref{section3}. Next, we evaluate NoPPA on eight tasks and report results in Section~\ref{section4}. The deeper analysis to NoPPA is described in Section~\ref{section5}. Finally, we summarize our work in Section~\ref{section6} and potential limitations in Section 7.

\section{Related Work}
\label{section2}
Recent studies in sentence embedding can be divided into two categories according to whether there are trainable parameters. 

There was a long history to capture sentence embedding using trainable models. Parametric methods usually train their language model with supervised tasks, and sentence embedding is an additional product. Most parametric models \citep{kiros2015skip, conneau2017supervised, logeswaran2018efficient, peters2018deep} are powered by recurrence neural networks (such as RNN, LSTM, and GRU) at the early stage. In contrast with the above methods, Sent2Vec \citep{pagliardini2017unsupervised} is an unsupervised method to learn sentence embedding with n-gram features. After 2017, different BERT-based methods \citep{cer2018universal, reimers2019sentence, wang2020sbert, li2020sentence, gao2021simcse, su2021whitening} were designed empowered by self-supervised learning from large-scale unlabeled corpora.

Non-parametric methods choose a more straightforward way that heavily relies on high-quality pre-trained word vectors \citep{wieting2015towards, mikolov2013efficient, pennington2014glove, joulin2016fasttext, salle2016matrix}. Individually weighted averaging each word embedding is the easiest way \citep{wieting2015towards, arora2017simple, ethayarajh2018unsupervised, yang2018parameter} based on the bag-of-words model. There were some attempts to remove the assumption of ignoring word orders from the bag-of-words models by capturing time information in the signal domain \citep{kayal2019eigensent, almarwani2019efficient}. Combining multi-resource word embedding \citep{mekala2016scdv, ruckle2018concatenated} and semantic subspace analysis \citep{wang2021efficient, ionescu2019vector} are methods of enhancing the original static word embedding. 

Normally, parametric models are expected to be better than non-parametric models. However, we need non-parametric methods in some scenarios where heavy computing requirements cannot be tolerated. Although the performance of recent non-parametric models has gradually improved, their computing complexity has also increased significantly, violating the purpose of designing them.

\section{Pairwise Attentive Random Walk Model for Sentence Embedding}
\label{section3}
The most significant limitation of bag-of-words models is treating each word equally across the whole corpus. This assumption is against human intuition that people change the meaning of a word respecting the surrounding context. Since the self-attention mechanism can capture surrounding word information well, combining the bag-of-words model and the attention mechanism should improve the vanilla bag-of-words model.

In this section, we first describe our pairwise attentive random walk model in Section~\ref{sec31}. Then we formalize the sentence embedding of the proposed language model in Section~\ref{sec32}. Our designs of non-parameters attention mechanisms to integrate contextual information is discussed in Section~\ref{sec33}. We cover the method used to remove the error introduced by the Taylor expansion in Section~\ref{sec34}. We finally analyze the time complexity in Section~\ref{sec35}. We summarize our method at Algorithm~\ref{algo1}.

\subsection{Pairwise Attentive Random Walk Model}
\begin{algorithm}[t]
\caption{NoPPA}
\label{algo1}
\begin{algorithmic}[1]
\Require Word embedding $\{v_w: w\in V\}$, Position embedding $\{pv_i:i \leq |S|\}$, Word probabilities $\{pr(w): w\in V\}$, A set of sentence $s \in S$, and Hyper-Parameters $a, k$.
\State {// Equation 8-12}
\For{sentence $s \in S$}
    \For{$i < |s|$}
        \State $v_w \gets \text{word vector for word }w$
        \State $v'_i \gets v_w + pv_i$
        \For{$j < |s|$}
            \State $A_{ij} \gets \text{Softmax}(\frac{v'_iv'^T_j}{\sqrt{d}})$
            \State $V_{ij} \gets \log_2(1 + (v'_i-v'_j)^2)$
        \EndFor
    
    \State $v'_w \gets \sum_{j=1}^nA_{ij}V_{ij}$
    \State $v'_w \gets \text{Concatenate}(v'_w, v_w)$
    
    \State $v_s \gets \frac{1}{|s|}\sum_{w\in{s}}\frac{a}{pr(w)+\frac{a}{2}}v'_w$
    \EndFor
\EndFor
\State

\State {// Equation 13 (do in training)}
\State $X \gets \text{form matrix } X \text{ which rows are }v_s$
    \State $U, S, V \gets \Call{SVD}{X}$
    \State $V_k \gets \text{choose the last }k\text{ rows of the matrix} V$
\State

\State {// Equation 14}
\For{sentence $s \in S$}
    \State $v_s \gets v_s - v_sV_k^\mathbf{T}V_k$
\EndFor
\Ensure Sentence embedding $v_s$ for $s \in S$
\end{algorithmic}
\end{algorithm}

\label{sec31}
The latent variable generative model \cite{arora2016latent} treats the corpus generation as a dynamic process. The process is driven by the random walk of a discourse vector $c_t\in \mathbb{R}^d$ at the time $t$, and each word $w$ in the vocabulary has a vector $v_w\in \mathbb{R}^d$. Both $c_t$ and $v_w$ are latent variables. The discourse vector represents the intent of the speaker. Thus, the probability of observing a word $w$ at time $t$ is:
\begin{equation}
\label{eq0}
    P(w|c_t)\propto Similarity(c_t, v_w).
\end{equation}
The discourse vector $c_t$ does a slow random walk during the generation so that a single discourse embedding $c_s$ can replace all the $c_t$ in the sentence $s=\{w_1, w_2, ..., w_n\}$ where $n$ is sentence length. 

In this work, we extend the random walk process from the uni-gram word to bi-gram word pairs because we notice that words could have different meanings in different contexts, and some words always appear together with the others to form phrases. Thus, we assume that each pair of words $w_i$ and $w_j$ in the vocabulary has an embedding $v_{ij}\in \mathbb{R}^d$ and define the contextual probability $Pc$ of observing a word $w_i$ in a sentence $s$ as:
\begin{equation}
    \label{eq1}
    Pc(w_i|c_s)= \sum_{j=1}^nA_{ij}D_{ij}(c_s),
\end{equation}
where
\begin{equation}
\label{eq2}
\begin{aligned}
D_{ij}(c_s) &= \frac{d(v_{ij}, c_s)}{Z_c(v_i)},\\
Z_c(v_i) &= \sum_{v_{j} \in |V|} d(v_{ij}, c_s).\\
\end{aligned}
\end{equation}
We first measure the similarity $D_{ij}(c_s)\in \mathbb{R}$ between the word pair embedding $v_{ij}$ and the discourse embedding $c_s$ to calculate the contextual probability $Pc(w_i|c_s)$. We also evaluate the probability of a word pair $A_{ij}\in [0,1]$ in line with language fluency. Then we have a condition that $\sum_{j=1}^nA_{ij}=1, i=1,2,...,n$ by assuming each word can only be triggered by one surrounding word. In this study, we use the angular distance between $v_{ij}$ and $c_s$ to measure their similarity. That is $d(v_{ij}, c_s) = 1 - \frac{\arccos \cos(v_{ij}, c_s)}{\pi}$.

To be more realistic, we consider the probability that a word appearing with a given discourse embedding $c_s$ is affected by the non-contextual/global probability $Pr(w_i)$ and the contextual/local probability $Pc(w_i|c_s)$. Thus, we measure the probability of observing a word $w_i$ as follows:
\begin{equation}
\label{eq3}
\begin{aligned}
P(w_i|c_s) &= \alpha Pr(w_i) + (1 - \alpha)Pc(w_i|c_s) \\
    &= \alpha Pr(w_i) + (1 - \alpha)\sum_{j=1}^nA_{ij}D_{ij}(c_s),     
\end{aligned}
\end{equation}
where $\alpha$ is a scalar and $Pr(w_i)$ is the uni-gram probability of word $w_i$ that appears in the corpus.

With Equation~\eqref{eq3}, we can define the probability of observing a sentence $s$ normalized with the sentence length $n$ as follows:
\begin{equation}
\label{eq4}
    P(s) = \sqrt[n]{\prod_{i=1}^nP(w_i|c_s)} \text{ .}
\end{equation}

\subsection{Sentence Embedding Estimation}
\label{sec32}
We treat the Maximum Log-Likelihood Estimation (MLE) of $c_s$ from Equation~\eqref{eq4} as the sentence embedding from the model. The log-likelihood of the sentence can be formalized as  
\begin{equation}
\begin{aligned}
\label{eq5}
F_s &= \frac{1}{n}\sum_{w_i \in s} F_{w_i} , \\
\end{aligned}
\end{equation}
where 
\begin{equation}
\label{eq6}
F_{w_i} = \log{[\alpha Pr(w_i) + (1-\alpha)\sum_{j=1}^nA_{ij}D_{ij}]} .
\end{equation}
Maximizing $F_s$ equals maximizing $F_{w_i}$. We can approximate $F_{w_i}$ using a first-degree Taylor expansion to simplify the calculation. We borrow the fundamental assumption proposed by \citet{arora2016latent} that the pair of words embedding $v_{ij}$ is roughly uniformly distributed in the latent space so that $Z_c$ can be seen as a constant. Thus, the first derivative of $F_{w_i}$ is
\begin{equation}
\begin{aligned}
\label{eq7}
\frac{\partial F_{w_i}}{\partial c_s} &=\frac{\sum_{j=1}^n\frac{A_{ij}\frac{\partial cos(v_{ij}, c_s)}{\partial c_s}}{\sqrt{1 - cos^2(v_{ij} c)}}}{\pi \exp(F_w(c))Z_c}, \\
\frac{\partial cos(v_{ij}, c)}{\partial c} &= \frac{v_{ij}}{||v_{ij}||_2 ||c_s||_2} - \frac{cos(v_{ij}, c_s)c_s}{||c_s||_2^2}.
\end{aligned}
\end{equation}
Assume that we can find a vector $v_0$ is orthogonal to any $v_{ij}$ with length $\frac{1}{||v_{ij}||}$. The approximation of $F_{w_i}$ on the vector $v_0$ is 
\begin{equation}
\begin{aligned}
\label{eq8}
F_{w_i}(c_s) &\approx  F_{w_i}'(v_0) + F_{w_i}'(v_0)c_s \\
            &\approx C + \frac{a}{\pi(Pr(w) + \frac{a}{2})}\sum_{j=1}^{n}A_{ij}v_{ij},
\end{aligned}
\end{equation}
where $C$ indicates a constant and $a = \frac{1-\alpha}{Z\alpha}$.

By applying MLE to estimate $c_s$ on Equation~\eqref{eq8}, we have 
\begin{equation}
\begin{aligned}
\tilde{c}_s &= \arg{max}_{c_s} \frac{1}{n}\sum_{w\in s}F_w(c_s) \\
          &\propto \frac{1}{n}\sum_{w_i\in s}\frac{a}{Pr(w_i) + \frac{a}{2}}\sum_{j=1}^{n}A_{ij}v_{ij}. \label{eq9} \\
\end{aligned}
\end{equation}

Equation~\eqref{eq9} indicates that the sentence embedding equivalents to weighted averaging of contextual embedding. Specifically, the contextual embedding is $\sum_{j=1}^{n}A_{ij}v_{ij}$, while the weight of each contextual embedding is $\frac{a}{Pr(w_i) + \frac{a}{2}}$. Comparing with SIF \citep{arora2017simple} and uSIF \citep{ethayarajh2018unsupervised} models, the primary significance of ours is using a pairwise bi-gram model instead of the uni-gram model. Moreover, we do not introduce a common discourse vector $c_0$ as they did.

\subsection{Contextual Embedding}
\label{sec33}
\label{ce}
The pairwise embedding $V_{ij}$ and the attention score $A_{ij}$ are keys to designing contextual embedding, while the $V_{ij}$ represents the meaning of the word pair $w_i$ and $w_j$, and the $A_{ij}$ measures how likely the word pair appears together. Our proposed method does not make any assumption to $V_{ij}$ and only one assumption to $A_{ij}$. Therefore, we have great freedom to define a variety of $V_{ij}$ and $A_{ij}$.

\subsubsection{Positional Word Embedding} 
To capture sequential information, we add the position embedding to the word embedding directly:
\begin{equation}
\begin{aligned}
v'_i = \text{WordEmbed}(w_i) + \text{PosEmbed}(i). \label{eq10}
\end{aligned}
\end{equation}
$PosEmbed(i)$ denotes the position embedding for the $i$th word in the sentence. We reference the definition of position embedding from the Transformer \citep{vaswani:17}:

\begin{equation}
\begin{aligned}
PosEmbed(i,2m)&=sin(\frac{pos}{10000^{2m/d_v}})\\
PosEmbed(i,2m+1)&=cos(\frac{pos}{10000^{2m/d_v}}), \label{eq11}\\
\end{aligned}  
\end{equation}
where $i$ is the word position, $m$ is the dimension being generated, and $d_v$ is the number of the entire position embedding dimension.

\subsubsection{Pairwise Embedding Using Log-Kernel}
It is tough to estimate pairwise embedding $v_{ij}$ accurately with a limited corpus because of the sparse latent space. For example, Wiki corpus has almost 180,000 unique tokens, and the total number of pairwise word embedding will be 32.4 billion. This sparse hidden parameter space makes any existing algorithms unable to obtain stable estimations. To avoid this issue, we design our pairwise word embedding as 

\begin{equation}
\begin{aligned}
v_{ij} = [v_i; \log_2(1 + (v'_j - v'_i)^2)], \label{eq12}
\end{aligned}
\end{equation}
where $v_i$ is the initial word embedding, and $[\text{ }\bigcdot\text{ ; }\bigcdot\text{ }]$ means the concatenation operation. We call this element-wise non-linear transform between $v'_j$ and $v'_i$ as the Log-Kernel.

\subsubsection{Non-Parametric Pairwise Attention}
Pairwise attention evaluates the probability of occurring a pair of words. The only assumption is $\sum_{j=1}^nA_{ij}=1$. The idea of pairwise attention is similar to the self-attention from Transformers, but our model is untrainable. An easy way to think of non-parametric features is the two words' relative position and semantic similarity. Thus, we directly dot product each two positional word embedding to measure the probability of a pair of words. We formalize our non-parametric pairwise attention as:

\begin{equation}
A_{ij} = softmax(\frac{v'_iv_{j}'^{\top}}{\sqrt{d}}),  
\label{eq13}
\end{equation}
where $d$ is the dimension of embedding $v'_i$.

\subsection{Noise Removal}
\label{sec34}
\label{nr}
We apply Taylor expansion to simplify the calculation of log-likelihood $F_w$ in Equation~\eqref{eq8}. However, the Taylor expansion reaches an approximate estimation, and some small error terms are ignored in Equation~\eqref{eq9}. Thus, the final sentence embedding removes the projections on singular vectors with the lowest singular values to remove these uncertainty introduced by the Taylor expansion.

We represent each sentence from the dataset with a vector and denote the matrix of sentence embedding as $X\in \mathbf{R}^{l\times d}$, where $l$ indicates the number of sentences and $d$ is the dimension of sentence vectors. Then, to apply singular value decomposition, we find three matrices $U \in \mathbf{R}^{l\times l}$, $S \in \mathbf{R}^{l\times d}$, and $V \in \mathbf{R}^{d\times d}$ that satisfy:
\begin{equation}
X = USV, \label{eq14}
\end{equation}
where $S$ is a diagonal matrix. 

We further assume that the last $k \leq d$ singular vectors with the smallest singular values contain the error terms of estimation. So the final estimation of each sentence vector will remove the projection on $V_k\in \mathbf{R}^{k\times d}$ as below

\begin{equation}
    \hat{c}_s = \tilde{c}_s - \tilde{c}_sV^{T}_kV_k. \label{eq15}
\end{equation}


\subsection{Time Complexity Analysis}
\label{sec35}
We denote the number of words in a sentence as $n$, the dimension of word embedding is $d$, and the number of noise singular vectors is $k$. Since we can pre-calculate $U$, $S$, and $V$ for a specific dataset during the training stage, inference sentence embedding time complexity for one sentence is $O(n^2d+k^2d)$. Since sentence length is roughly 20 to 30 words, and the sufficient $k$ is always under 20, NoPPA should have lower time complexity than solutions based on latent space analysis \citep{yang2018parameter, wang2021efficient}.

\begin{table*}[htbp]
\footnotesize
\begin{center}
\begin{tabular}{lccccccccc}
\toprule
Model                 & MR   & CR   & SUBJ & MPQA & SST2 & TREC & MRPC & SICK-E & Avg \\
\hline
FastSent            & 70.8 & 78.4 & 88.7 & 80.6 &      & 76.8 & 72.2 &        & 77.9   \\
USE (DAN)             & 74.0 & 80.5 & 91.9 & 83.5 & 80.3 & 89.6 & 71.8 & 80.4   & 81.5   \\
Sent2Vec                & 75.8 & 80.3 & 91.1 & 85.9 &      & 86.4 & 72.5 &        & 82.0   \\
SBERT               & 83.6 & 89.4 & 94.4 & 89.9 & 89.0 & 89.6 & 76.0 && 87.4   \\
\hline
GloVe-avg*            & 77.4   & 80.0   & 91.6 & 87.8 & 82.2 & 84.0 & 73.2 & 79.2 & 81.9  \\
SIF*                   & 77.6   & 78.7   & 91.3 & 87.3 & 82.3 & 79.6 & 71.9 & 75.1 & 80.5  \\
TFIDF*                  & 77.5   & 79.5   & 91.9 & 87.7 & 81.7 & 83.6 & 74.0 & 79.3 & 81.9 \\

VLAWE                 & 77.7 & 79.2 & 91.7 & 88.1 & 80.8 & 87.0 & 72.8 & 81.2 & 82.3   \\
DCT*                    & 78.3   & 80.0   & 92.5 & 88.2 & 82.5 & 88.6 & 73.6 & 81.8 & 83.2  \\
S3E*           & 77.9 & 79.6 & 91.5 & 87.0 & 82.6 & 83.8 & 73.9 &  78.4  &  81.8 \\
GEM              & 78.8 & 81.1 & 93.1 & 89.4 & 83.6 & 88.6 & 73.4 & 85.3   & 84.2   \\

\hline

CE-avg          & 77.4 & 80.1 & 92.4 & 88.0 & 81.9 & 87.2 & 73.6 & 81.5 & 82.7  \\
CE-avg+NR       & 77.7 & 80.4 & 92.4 & 88.1 & 83.4 & 87.2 & 73.6 & 81.5 & 83.1  \\
CE+SFW      & 77.9 & 80.2 & 92.4 & 88.2 & 82.9 & 88.0 & 74.1 & 81.2 & 83.2  \\
NoPPA            & $78.0_{\pm0.12}$  & $80.5_{\pm0.15}$  & $92.9_{\pm0.06}$  & $88.4_{\pm0.03}$  & $84.1_{\pm0.12}$ & $88.2_{\pm0.22}$ & $74.7_{\pm0.21}$ & $81.6_{\pm0.16}$ & $83.6$ \\
\bottomrule
\end{tabular}

\caption{Sentence embedding performance on downstream tasks.} \label{tb1}
\end{center}
\end{table*}

\section{Experiment}
\label{section4}
We evaluate our method on eight different downstream tasks. We also conduct ablation experiments to discuss the sufficiency of each strategy of our approach. The sensitivity analysis finally discovers the impact of hyper-parameters. 

\subsection{Benchmarks}
We not only compare against other bag-of-words-based methods, including \textbf{Average Embedding}, \textbf{TF-IDF Weighted Average}, and \textbf{SIF} \citep{arora2017simple}, but also with improved methods such as \textbf{DCT} \citep{almarwani2019efficient}, \textbf{VLAWE} \citep{ionescu2019vector}, \textbf{GEM} \citep{yang2018parameter}, and \textbf{S3E} \citep{wang2021efficient}. We further compare our method with popular parametric methods with similar sentence embedding dimensions such as \textbf{FastSent} \citep{hill2016learning}, \textbf{Sen2Vec} \citep{moghadasi2020sent2vec}, \textbf{USE(DAN)} \citep{cer2018universal}, and \textbf{SBERT} \citep{reimers2019sentence}. 

\subsection{Tasks}
We evaluate our method on eight classification tasks. These tasks consist of fine-grained sentiment classification (MR, CR, MPQA, SST-2) \citep{pang2005seeing, hu2004mining, wiebe2005annotating,socher2013recursive}, question type classification (TREC) \citep{voorhees2003overview}, subjectivity/objectivity classification (SUBJ) \citep{pang2004sentimental}, entailment relation classification (SICK-E) \citep{lai2014illinois}, and paraphrase identification (MRPC) \citep{dolan2004unsupervised}. These datasets can measure how well the sentence embedding is.

\subsection{Detail Settings}
We evaluate our methods with the SemEval toolkit \citep{conneau2018senteval}, a library for measuring the quality of sentence embedding. For tasks, we build a one-hidden-layer MLP with 50 parameters as a classifier. The classifier is optimized with Adam \citep{kingma2014adam} with 64 batch sizes and a 0.0 dropout rate. We tune hyper-parameters $k$\footnote{$k\in [0, 24]$}, and $a$\footnote{$a\in \{0.01, 0.15\}$} with the Bayesian optimization over 40 times. Our model relies on GloVe \citep{pennington2014glove} as static word embedding and word frequencies collected from Wiki corpus. To fairly compare methods, we testify some benchmarks by ourselves with the same experimental settings.

\subsection{Main Results}
Experimental results on eight different supervised downstream tasks are listed at Table~\ref{tb1}. We report both mean and standard deviation scores for the full model NoPPA over five different random seeds in the last line. The models in the table marked with * are testified with the same classifier setting by ourselves, and we also apply grid search to find out the best hyper-parameters for them. All non-parametric methods reported in Table~\ref{tb1} use GloVe word embedding only. We also show ablation study results on this table which will be discussed later. Here, CE denotes averaging contextual embedding followed by Section~\ref{ce}, NR means noisy removal strategy described in Section~\ref{nr}, SFW stands for the smooth frequency weight $\frac{a}{Pr(w_i)+\frac{a}{2}}$ from Equation~\eqref{eq9}, CE-avg indicates SFW constantly equals 1 so that we can evaluate the quality of contextual embedding only.

\paragraph{NoPPA generates high-quality sentence embedding for downstream tasks.}
The performance of our method on all downstream tasks is shown in the last line of Table~\ref{tb1}. Our method makes significant progress compared to all weighted averaging-based methods, including AVG, SIF, and TFIDF. We also outperform DCT, which is designed to capture sequential information using discrete cosine transform, on average for eight datasets. We reach better performance than most latent space analysis methods including VLAWE and S3E. Although we didn't get the GEM score, it's worth mentioning that 
its time complexity is significantly higher than ours since it uses SVD in the inference phase. Compared with parametric models, we do better than most models, except SBERT-WK.

\paragraph{CE, SFW, and NR are sufficient strategies.}
Empirical speaking, the architecture of NoPPA can be separated into three components, which are contextual embedding in Section~\ref{ce}, smooth frequency weight $\frac{a}{Pr(w_i) + \frac{a}{2}}$, and noisy removal strategy in Section~\ref{nr}. According to Table~\ref{tb1}, CE-avg constantly makes positive contributions to improve the baseline model GloVe-avg among all datasets except MR. Both SFW and NR make contributions to improve NoPPA on average. 

\paragraph{NoPPA is slowed down by Log-Kernel.}
S3E \citep{wang2021efficient} is the fastest method among all benchmarks. We compare the inference time between NoPPA and S3E on 19,854 sentences from the SICK-E dataset. The total inference time of S3E on our environment\footnote{Intel i9-11900KF @ 3.5GHz} is 3.23\footnote{10 times average is 3.2267, standard error is 0.0249} seconds, while that of NoPPA is 7.55\footnote{10 times average is 7.5489, standard error is 0.0738} seconds. However, we found that removing the $\log_2$ operation will only take 2.45\footnote{10 times average is 2.4524, standard error is 0.0259} seconds, 1.32 times faster than S3E. This result corroborates our time complexity analysis results to NoPPA in Section 3 ignoring implementation differences. Since the run time speed of NoPPA will be affected by different implementations of the $\log_2$ function, we suggest users who seek higher speed explore other non-linear kernels.

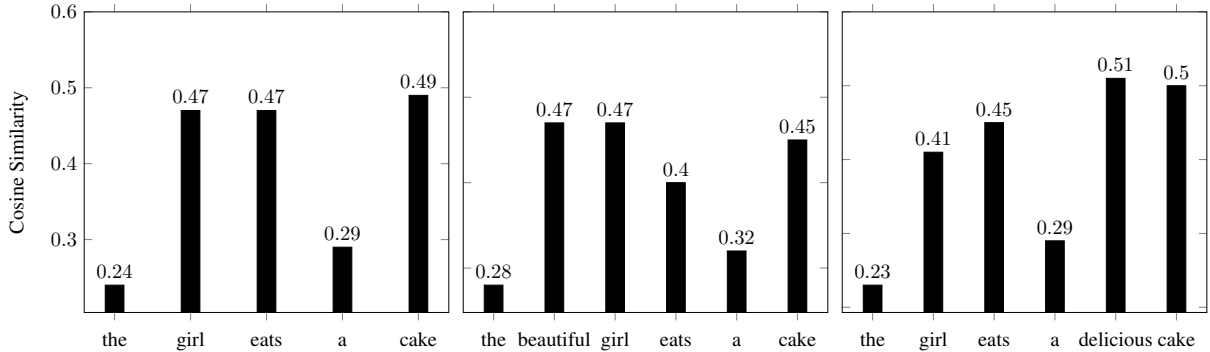
\begin{figure*}[htbp]
\centering
\begin{tikzpicture}[scale=0.7, transform shape]
\begin{axis} [
    ybar,
    symbolic x coords={the,girl,eats,a,cake},
    xtick={the,girl,eats,a,cake},  
    xticklabel style={text height=2ex}, 
    ylabel=Cosine Similarity,
    anchor=left of west,
    ymax=0.6,
    nodes near coords,
    nodes near coords align={vertical},  
    name = left,
]
\addplot[ybar,fill] coordinates{(the,0.24) (girl,0.47) (eats,0.47) (a,0.29) (cake,0.49)};
\end{axis}

\begin{axis} [
    ybar,
    yticklabel={\ },
    symbolic x coords={the,beautiful,girl,eats,a,cake},
    xtick={the,beautiful,girl,eats,a,cake},  
    xticklabel style={text height=2ex}, 
    anchor=left of west,
    ymax=0.6,
    nodes near coords,
    nodes near coords align={vertical},  
    name = mid,
    at=(left.right of east),
]
\addplot[ybar,fill] coordinates{(the,0.28) (beautiful,0.47) (girl,0.47) (eats,0.40) (a,0.32) (cake,0.45)};
\end{axis}

\begin{axis} [
    ybar,
    yticklabel={\ },
    symbolic x coords={the,girl,eats,a,delicious,cake},
    xtick={the,girl,eats,a,delicious,cake},  
    xticklabel style={text height=2ex}, 
    anchor=left of west,
    ymax=0.6,
    nodes near coords,
    nodes near coords align={vertical},  
    name = right,
    at=(mid.right of east),
]
\addplot[ybar,fill] coordinates{(the,0.23) (girl,0.41) (eats,0.45) (a,0.29) (delicious,0.51) (cake,0.50)};
\end{axis}
\end{tikzpicture}
\caption{Contributions of each word to their sentence embedding.}
\label{fig2}
\end{figure*}

\section{Analysis}
\label{section5}
This section answers two questions:
\begin{enumerate}
\item How to choose the best hyper-parameter $a$ for different datasets?
\item Why does the model work?
\end{enumerate}

\subsection{Impacts of hyper-parameter $a$}
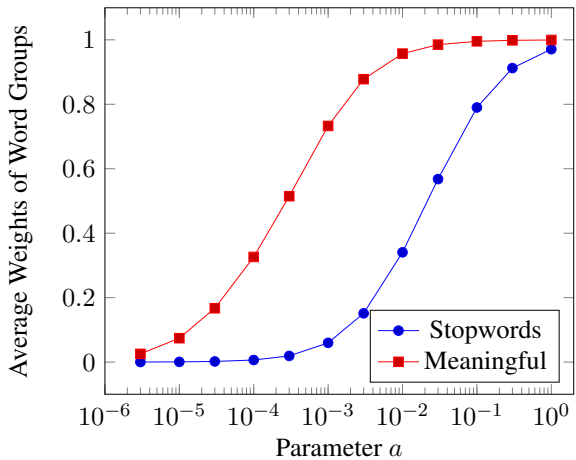
\begin{figure}[]
\centering
\begin{tikzpicture}[scale=0.9, transform shape]
\begin{axis} [
    xmode=log,
    xlabel=Parameter $a$,
    ylabel=Average Weights of Word Groups,
    xmin = 0.000001,  
    xmax = 2, 
    legend pos=south east,
]
 \legend {Stopwords, Meaningful};
\addplot coordinates {(1.0,0.970877) (0.3,0.912303) (0.1,0.790042) (0.03,0.567890) (0.01,0.340638) (0.003,0.151312) (0.001,0.059712) (0.0003,0.019229) (0.0001,0.006551) (0.00003,0.001981) (0.00001,0.000662) (0.000003,0.000199)};
\addplot coordinates {(1.0,0.9995) (0.3,0.9985) (0.1,0.9954) (0.03,0.9850) (0.01,0.9572) (0.003,0.8779) (0.001,0.7328) (0.0003,0.5148) (0.0001,0.3263) (0.00003,0.1671) (0.00001,0.0742) (0.000003,0.0256)};
\end{axis}
\end{tikzpicture}
\caption{Impacts of hyper-parameter $a$ to word weights.}
\label{fig1}
\end{figure}

Since we count word frequency $Pr(w_i)$ from large corpus instead of downstream datasets, smooth frequency weight $weight(w_i)=\frac{a}{Pr(w_i) + \frac{a}{2}}$ has no change by giving a specific $a$ over all datasets. To analyze how parameter $a$ influences the weights of words, we calculate the average weights of 13 stop words\footnote{\{of, the, a, in, at, to, with, by, and, are, is, ".", ","\}} and 16 meaningful words\footnote{\{film, man, women, dogs, cats, name, air, phone, special, large, past, emotional, easy, need, found, show\}} by setting different $a$. Then, we plot the average weights of the different groups in Figure~\ref{fig1}.

According to Figure~\ref{fig1}, the average weight of stopwords drops faster than that of meaningful words when we decrease parameter $a$ gradually. That makes it possible to filter stop words and enhance the contributions of meaningful words by setting different $a$. From Figure~\ref{fig1}, we conclude that the best $a$ to distinguish the two groups of words should be between 1e-1 and 1e-2. If $a$ is smaller than $10^{-2}$, then the meaningful words will be weakened. The experiments have supported this assertion that NoPPA never reach their best score with $a$ higher than $0.1$ on all datasets (see Appendix~\ref{sec:appendixa}). 

\subsection{What Knowledge Does NoPPA Learn}
We try to understand the model from different views. One of the views is a result-oriented approach which means studying the sentence embedding generated by the method. The second view is a process-oriented approach in which we examine the model to understand what it learned. 

\paragraph{Understanding Context Topics}
We take three sentences\footnote{Three sample sentences are: "the girl eats a cake", "the beautiful girl eats a cake", "the girl eats a delicious cake".} as examples and compare the contribution of the same word among different sentences. In detail, we concatenate the original word embedding of each word horizontally so that the size of each word embedding is the same as the size of sentence embedding. Then we draw the cosine similarity between the concatenated word embedding and the sentence embedding in Figure~\ref{fig2}.

The three sentences use the simplest syntax: a subject, a verb, and an object. Everyone can easily predict the topics of these sentences when they read through them. For example, in the second statement, we might infer that the speaker will talk about how beautiful the girl is in the following conversation. People always pay more attention to the words included in their predicted topics. It means that the subject ("girl") should make more contribution than the object ("cake") to the sentence meaning in the second sentence.

We observe these dynamic weights that change with contextual topics in Figure~\ref{fig2}. It indicates that our model dynamically assigns different attention to the same word regarding different contexts. Particularly, the subject ("girl") in the first and the second sentence has 47\% similarity to the sentences. In contrast, it has only 41\% similarity in the third sentence. We can find this phenomenon in analyzing the verb ("eats") and the object ("cake") as well. Surprisingly, although stopwords ("the", "a") show this phenomenon as well, they are consistently assigned with low attention. Thus, we conclude that our method can detect contextual topics and give more attention to topic words.

\paragraph{Detecting Linguistic Phrases and Causation}
\begin{figure}[htp]
\centering
\includegraphics[scale=0.55]{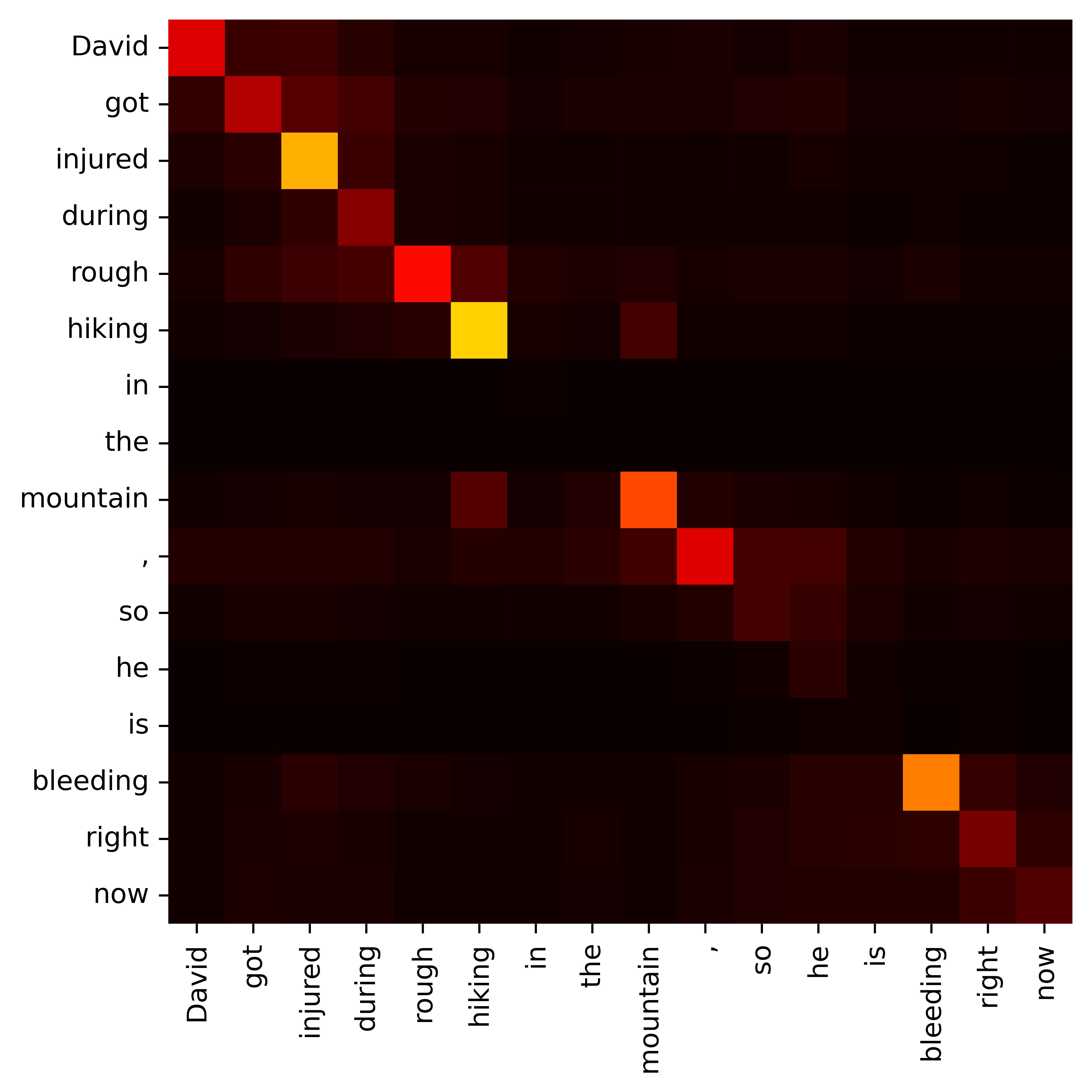}
\caption{Non-Parametric Self-Attention Heatmap.}
\label{fig3}
\end{figure}
In linguistic, sentence meaning has hierarchical levels. A robust language model can detect sophisticated semantic relationships. To investigate what kinds of knowledge the NoPPA uses, we choose a sentence \footnote{"David got injured during rough hiking in the mountain, so he is bleeding right now"} and draw the heat map for the pairwise weight score from the NoPPA in Figure~\ref{fig3}. 

The example sentence contains hierarchical semantic relationships. In the lexical level, it has common linguistic phrases such as "got injured", "rough hiking in somewhere", "right now". The causal relationship between "injured" and "bleeding" is the next level of semantic understanding. Among them, the top-level semantic meaning is coreference resolution between "David" and "he". 

For the first scan to Figure~\ref{fig3}, we can easily find that the model captures all kinds of linguistic phrases we mentioned before. Nevertheless,  looking at Figure~\ref{fig3} in detail, we will surprisingly notice that the line of "bleeding" slightly lights up the column "injured" as well. Actually, "bleeding" assigns a weight of 4.8\% to "injured", which is the third-highest weight assigned by "bleeding"\footnote{The attention score of "bleeding" to other words are David=1.21\%, got=2.17\%, injured=4.8\%, during=3.41\%,  rough=2.51\%, hiking=1.81\%, in=1.49\%, the=1.55\%, mountain=1.21\%, ","=2.02\%, so=2.68\%, he=4.59\%, is=4.51\%, bleeding=55.25\%, right=6.47\%, now=3.5\%. Sum of all attention score of "bleeding" is weighted by SFW to 99.19\%.}. The two highest weights are itself and the word "right" next to it. According to this finding, although the attention score is not as large as expected, we can still conclude that NoPPA can identify long-range causal inference relationships. However, the problem that NoPPA cannot handle the coreference resolution well still remains. We provide more examples in Appendix~\ref{sec:appendixb}.

\section{Conclusion}
\label{section6}
We propose Non-Parametric Pairwise Attention Random Walk Model (NoPPA) to generate high-quality sentence embedding with a low computing complexity. NoPPA first constructs contextual embedding (CE) to capture the contextual information for each word with pre-trained static word embedding, pre-computed static position embedding, and an element-wise non-linear transform. Then, pre-counted word frequency is applied to assign non-contextual weights (Smooth Frequency Weight, SFW). Next, we weight average contextual embedding based on the assigned SFW. Finally, the projections on the last few principal components are subtracted to remove the estimating errors (Noise Removal, NR). NoPPA is a non-parametric method, and it runs in only $O(n^2d+k^2d)$ time complexity during the inference stage. We evaluate NoPPA on eight downstream text classification datasets. According to the results, NoPPA constantly outperforms all bag-of-words-based methods and does better than non-parametric methods using time information and most of the latent-space-analysis-based methods on average with lower time complexity. Visualization analysis supports that NoPPA can detect context topics, common phrases, and long-range word-word causation.

\section*{Limitations}
\label{section7}
First, NoPPA sentence embedding remains most properties of the word embedding it uses since NoPPA is simple. Thus, if the word embedding is anisotropic \citep{reimers2019sentence}, the NoPPA sentence embedding will be anisotropic. In other words, NoPPA may not be suitable for information retrieval systems before doing post-processing such as whitening \citep{huang2021whiteningbert}.

Second, solving $log_2(\cdot)$ used as non-linear function relies on complex algorithms in all numerical computing libraries. Thus, future research can explore other faster non-linear kernels.

\clearpage
\bibliography{anthology,custom}
\bibliographystyle{acl_natbib}

\clearpage
\appendix
\section{More Visualization Examples}
\label{sec:appendixb}
We provide two more examples of non-parametric self-attention results to illustrate how NoPPA utilizes common phrases and word-level causality.

\begin{figure}[htp]
\centering
\includegraphics[scale=0.5]{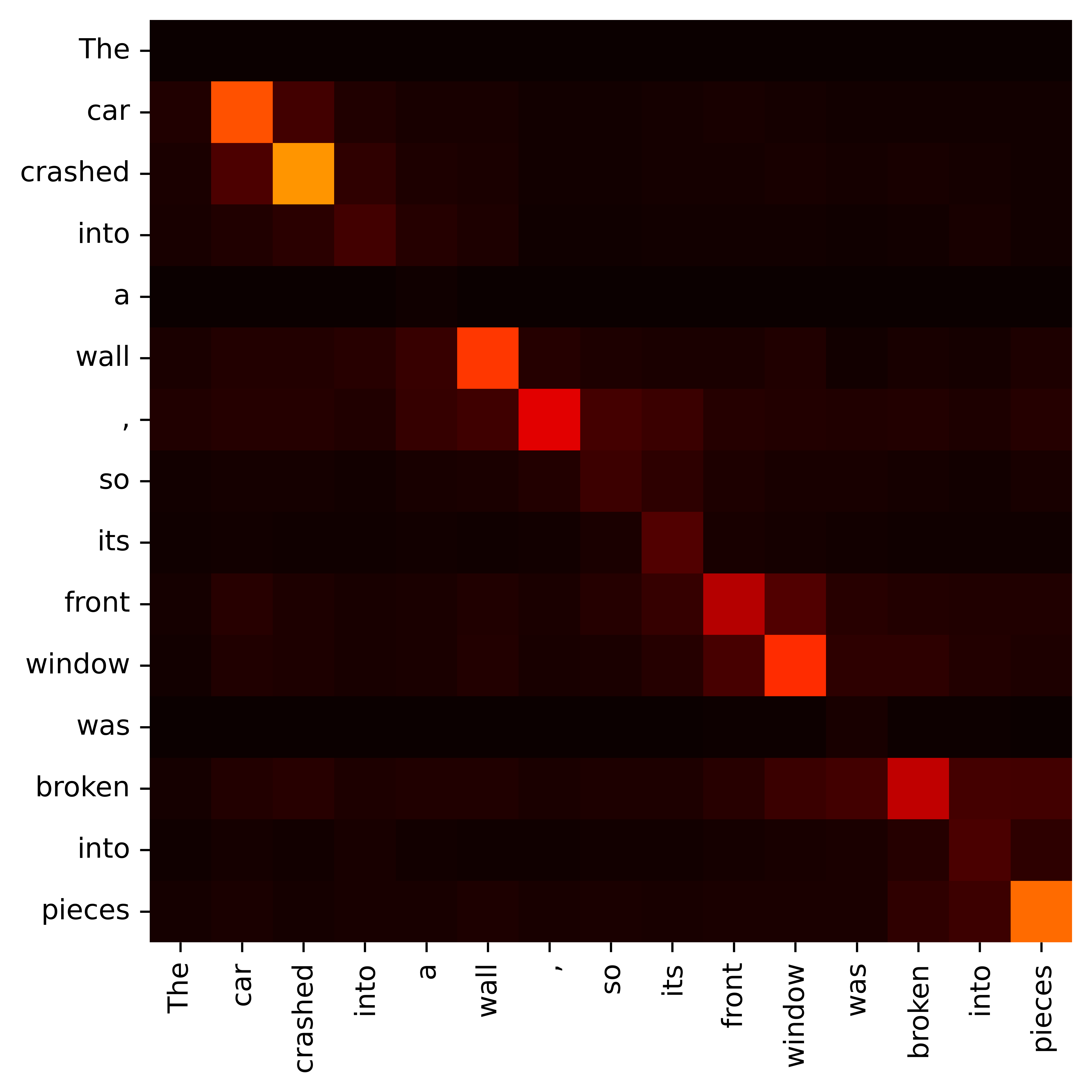}
\caption{Additional Examples 1 to Non-Parametric Self-Attention Heatmap.}
\label{fig4}
\end{figure}

In Figure~\ref{fig4}, NoPPA finds out "car crashed into", "front window", and "broken into pieces", which are widely used in daily conversations. NoPPA also detects the causation between the word "crashed" and the word "broken" with a score of 4.65\%, while the other words with higher weight scores are "window was broken into pieces".

\begin{figure}[htp]
\centering
\includegraphics[scale=0.5]{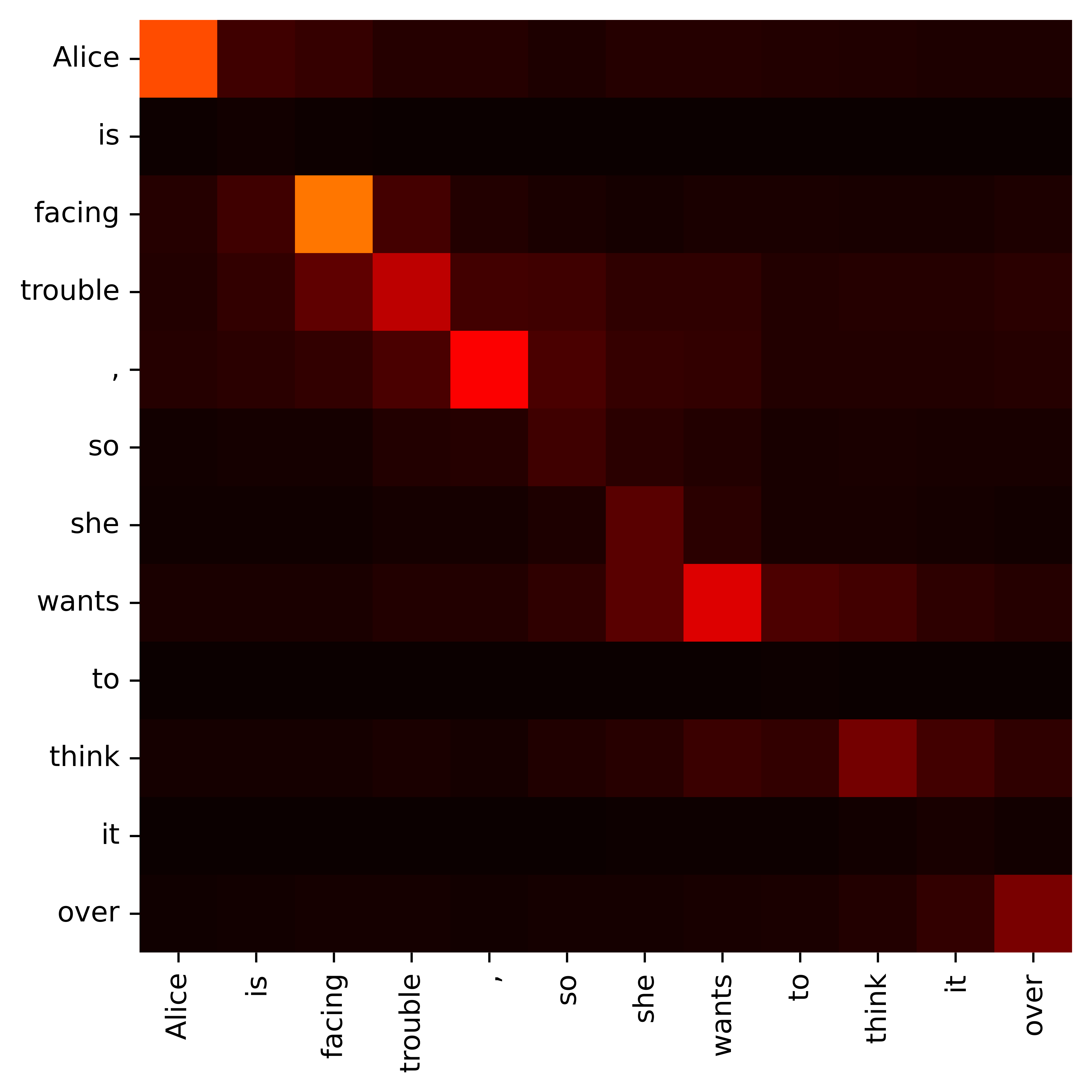}
\caption{Additional Examples 2 to Non-Parametric Self-Attention Heatmap.}
\label{fig5}
\end{figure}
In Figure~\ref{fig5}, NoPPA finds out the two common phrases "face trouble" and "think it over", as well as the causation between the word "trouble" and the word "think". More precisely, the word "think" assigns 2.62\% weight to the word "trouble". This is the highest score for all five words in the first part of the sentence.

\section{Hyper-Parameters for Each Dataset}
\label{sec:appendixa}
We record our bayesian search results to the best hyper-parameter setting for each dataset in Table~\ref{tb2}.
\begin{table}[htbp]
\footnotesize
\begin{center}

\begin{tabular}{clcccc}
\toprule
Seed & Dataset               & $a$ & $k$ & test-acc \\
\hline
\multirow{8}*{1034} & MR & 0.05 & 22 & 78.27 \\
 & SST2 & 0.1 & 2 & 84.29 \\
 & SUBJ & 0.03 & 21 & 92.90\\
 & MPQA & 0.1 & 6 & 88.35 \\
 & CR & 0.1 & 11 & 80.43 \\
 & SICK-E & 0.03 & 21 & 81.83 \\
 & MRPC & 0.1 & 6 & 74.84 \\
 & TREC & 0.1 & 16 & 88.00 \\
 \hline
\multirow{8}*{1314} & MR & 0.04 & 15 & 77.96 \\
 & SST2 & 0.1 & 4 & 84.13 \\
 & SUBJ & 0.02 & 9 & 92.91\\
 & MPQA & 0.01 & 5 & 88.38 \\
 & CR & 0.04 & 15 & 80.34 \\
 & SICK-E & 0.01 & 9 & 81.55 \\
 & MRPC & 0.1 & 1 & 74.67 \\
 & TREC & 0.06 & 1 & 88.20 \\
 \hline
\multirow{8}*{20220505} & MR & 0.01 & 15 & 78.04 \\
 & SST2 & 0.1 & 5 & 83.91 \\
 & SUBJ & 0.03 & 7 & 92.8\\
 & MPQA & 0.02 & 3 & 88.31 \\
 & CR & 0.02 & 15 & 80.72 \\
 & SICK-E & 0.01 & 14 & 81.57 \\
 & MRPC & 0.01 & 5 & 74.55 \\
 & TREC & 0.1 & 7 & 88.0 \\
 \hline
\multirow{8}*{20220508} & MR & 0.1 & 12 & 77.93 \\
 & SST2 & 0.07 & 2 & 84.07 \\
 & SUBJ & 0.02 & 5 & 92.84\\
 & MPQA & 0.03 & 11 & 88.39 \\
 & CR & 0.09 & 2 & 80.53 \\
 & SICK-E & 0.03 & 2 & 81.71 \\
 & MRPC & 0.02 & 2 & 74.38 \\
 & TREC & 0.03 & 13 & 88.20 \\
 \hline
\multirow{8}*{20220904} & MR & 0.02 & 12 & 77.99 \\
 & SST2 & 0.07 & 6 & 84.13 \\
 & SUBJ & 0.08 & 12 & 92.96\\
 & MPQA & 0.1 & 3 & 88.4 \\
 & CR & 0.1 & 14 & 80.61 \\
 & SICK-E & 0.07 & 22 & 81.37 \\
 & MRPC & 0.1 & 14 & 74.96 \\
 & TREC & 0.06 & 17 & 88.60 \\
\bottomrule
\end{tabular}
\caption{Best hyper-parameters for different datasets}\label{tb2}
\end{center}
\end{table}

\end{document}